\documentclass[runningheads]{llncs}

\usepackage{eccv}

\usepackage{eccvabbrv}

\usepackage{graphicx}
\usepackage{booktabs}
\usepackage[table]{xcolor}
\usepackage{array}

\usepackage[accsupp]{axessibility}  

\usepackage{hyperref}

\usepackage{orcidlink}
\usepackage{multirow}

\usepackage{algorithm}
\usepackage{algorithmic}
\usepackage{amsmath}
\usepackage{marvosym}

\begin{document}
\title{GeMoE: Gating Entropy is All You Need for Uncertainty-aware Adaptive Routing in MoE-based Large Vision-Language Models}

\titlerunning{GeMoE}

\newcommand{\corrauth}{\textsuperscript{\textrm{\Letter}}}

\author{
    Chaoxiang Cai\inst{1}$^{\dag}$ \and
    Minghe Weng\inst{2}$^{\dag}$ \and
    Jie Li\inst{2} \and
    Yibo Jiang\inst{1} \and
    Longrong Yang\inst{2} \and
    Zequn Qin\inst{1} \and
    Xi Li\inst{2}$^{\textrm{\Letter}}$
}

\authorrunning{C. Cai et al.}

\institute{
    School of Software Technology, Zhejiang University \and
    College of Computer Science and Technology, Zhejiang University \\
    \email{\{cxcai,wengminghe,li.jie,jiangyibo,longrongyang,qinzequn, xilizju\}@zju.edu.cn} \\
    \begingroup
        \renewcommand{\thefootnote}{\dag} \footnotetext[1]{Equal contribution.}
        \renewcommand{\thefootnote}{\textrm{\Letter}} \footnotetext[2]{Corresponding author.}
    \endgroup
}

\maketitle

\begin{abstract}
    With the increase in model parameters and training data, the instruction following and generalization capabilities of Large Vision-Language Models (LVLMs) have been significantly improved.
    Based on the Mixture-of-Experts (MoE) architecture, LVLMs expand their parameter capacity while maintaining the inference cost.
    However, traditional MoE methods employ a Top-$k$ static routing strategy, which fails to account for variations in the input and adaptively select the number of experts, resulting in suboptimal resource utilization.
    In this paper, we propose viewing token routing as an information encoding task, framing dynamic routing as a Minimum Description Length (MDL) problem in encoding.
    By validating the connection between MDL and gating entropy in the MoE scenario, we introduce Gating Entropy-based Uncertainty-aware Adaptive Routing (GeMoE) for MoE.    
    Unlike traditional static or heuristic-based dynamic routing methods, GeMoE explicitly models the trade-off between model complexity and performance.
    By using gating entropy to assess the complexity of tokens, GeMoE adaptively determines the number of experts each token should engage.
    On a wide range of backbones and benchmarks, our method achieves \textbf{\textit{99.5\%}} average performance retention compared to the original static routing, while improving average expert activation sparsity by \textbf{\textit{36.5\%}}.
    The code will be publicly available at \url{https://github.com/caichaoxiang/GeMoE}.
\end{abstract}

\section{Introduction}
\label{sec:intro}

The rapid growth of model parameters and training data has enabled Large Vision–Language Models (LVLMs) to achieve significant improvements across a wide range of tasks~\cite{liu2024visual, chen2024internvl, bai2023qwen, zhang2023llavar, wu2024deepseek}.
However, as these models continue to scale up, their size and resource demands present substantial challenges for deployment and real-world application.
In response to these issues, the Mixture-of-Experts (MoE) architecture has emerged as a widely adopted solution~\cite{jacobs1991moe, shazeer2017outrageously}. MoE replaces traditional Feed-forward Network (FFN) layers with a set of specialized experts, activating only a subset of them for each input.
This approach allows the model to expand its capacity without a proportional increase in inference computation, making it an appealing solution for scaling large models~\cite{dai2024deepseekmoe, qwenmoe}.

\begin{figure}[t]
    \centering
    \includegraphics[width=0.98\linewidth]{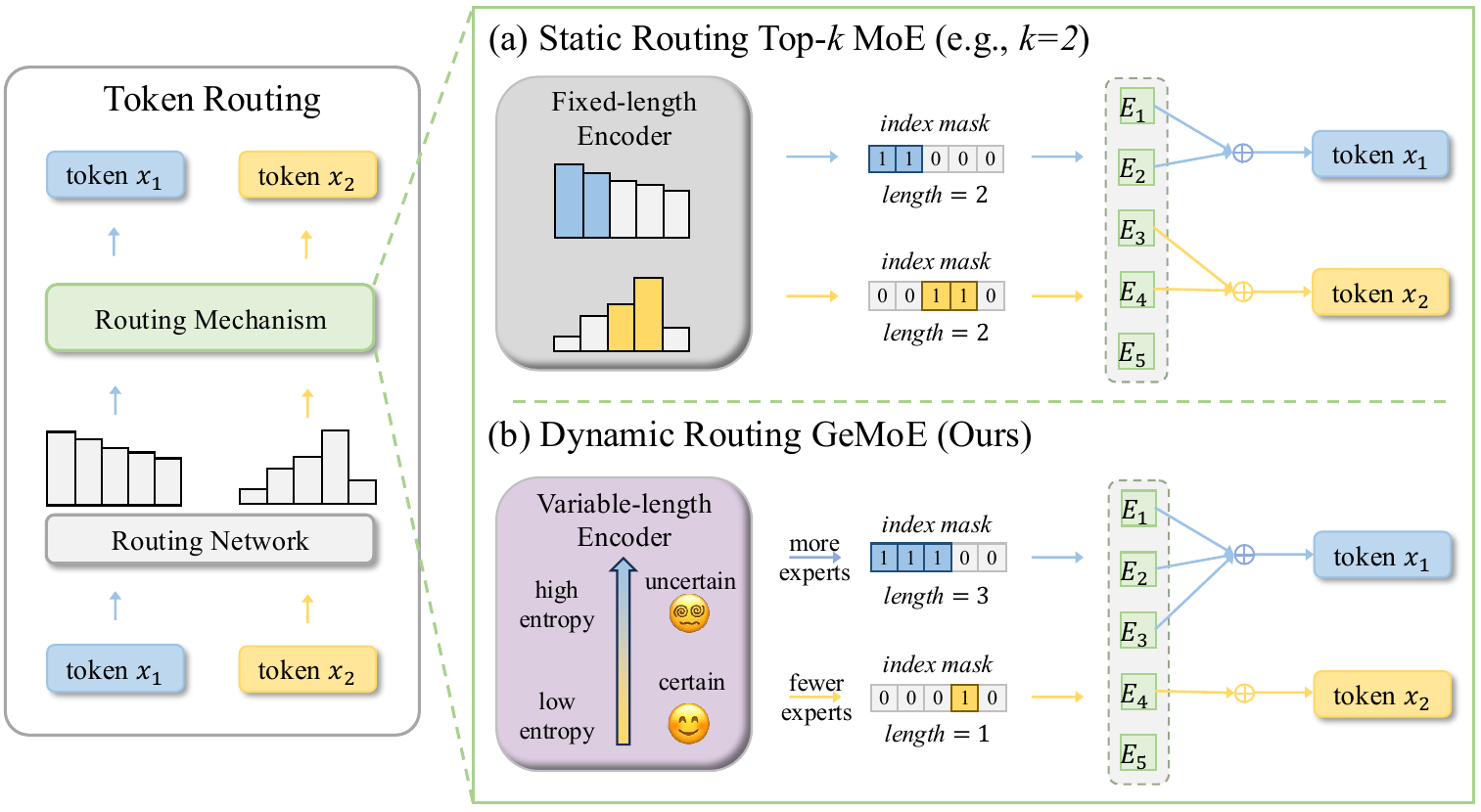}
    \caption{Comparison of static routing Top-$k$ MoE and our dynamic routing GeMoE: (a) Top-$k$ assigns a fixed number of experts to each token. (b) GeMoE dynamically determines the number of experts to activate based on the token's gating entropy. Gating entropy reflects the token's uncertainty in expert selecting. Higher entropy indicates greater uncertainty, requiring more experts to be selected.}
    \label{fig_intro}
\end{figure}

The core of MoE is token routing, which can be viewed as an information encoding task.
In this process, the set of experts acts as a codebook, with each expert representing a codeword.
Token routing's goal is to select the appropriate codewords to encode the tokens.
Traditional static Top-$k$ routing as shown in Fig.~\ref{fig_intro} (a) uses as a fixed-length encoder, assuming uniform information distribution across tokens and allocating equal expert resources to each token.
This leads to inefficiencies, as low-information tokens waste computational resources, while high-information tokens are under-resourced and lose semantic content.

Recent dynamic routing methods, such as pseudo-expert construction~\cite{zeng2024adamoe, jin2025moeplusplus} and expert threshold learning~\cite{guo2025dynamic, wang2025remoe, huang2024harder, yue2025adak}, aim to improve both resource efficiency and model performance.
As shown in Fig.~\ref{fig_intro} (b), dynamic routing can be viewed as a variable-length encoding method, optimizing the Minimum Description Length (MDL)—the total length required to encode both the model and its data fit.
The goal of optimal dynamic routing is to minimize the MDL.
However, current methods are based on heuristics and fail to explicitly balance model complexity with performance, resulting in inefficient expert allocation.

To this end, we propose Gating Entropy-based Uncertainty-aware Adaptive Routing for MoE (GeMoE), which incorporates two key design principles:
($i$) \textbf{Looking for a proxy to decrease the MDL}.
Since MDL is difficult to compute directly, it is necessary to introduce a proxy.
In this work, we use gating entropy as the proxy to reduce the number of experts assigned to tokens with low gating entropy, thus minimizing the MDL.
Specifically, when adding more experts improves the model’s ability to fit the data (defined as information gain) beyond a certain threshold, the addition of experts helps reduce the MDL.
Higher gating entropy indicates a more evenly distributed expert capacity, and in such cases, adding experts provides a higher information gain.
Therefore, more experts should be assigned to tokens with high gating entropy, while fewer experts should be allocated to tokens with low gating entropy.
This is further validated by the analysis in Section.~\ref{sec:method}.
($ii$) \textbf{Expert assignment prediction based on gating entropy}.
We introduce an Expert Assignment Predictor (EAP) that maps gating entropy to the required number of experts.
EAP takes token features as input and outputs the corresponding expert count for each token.
We design a monotonic loss function to enforce a positive correlation between the number of experts required by a token and its gating entropy, ensuring that tokens with higher entropy are assigned more experts.

In conclusion, our contribution can be summarized as:

\begin{itemize}
    \item We view token routing from an information encoding perspective, and frame dynamic routing as a MDL problem.
    \item By linking MDL and token gating entropy through information gain, we demonstrate that when a token has high gating entropy, adding an expert for that token provides significant information gain, thus reducing the MDL.
    \item We propose a gating entropy-based adaptive routing method, GeMoE, predicting the number of experts. Extensive experiments across backbones and benchmarks compared to the original static routing show that our method achieves a good trade-off between utilization and performance.
\end{itemize}

\section{Related Works}
\label{sec:related}

\subsection{Large Vision-Language Models}

Building on the success of Large Language Models (LLMs)~\cite{yang2025qwen3, comanici2025gemini}, recent research has expanded their capabilities to LVLMs for real-world visual applications.
Frameworks like DeepSeek-VL2~\cite{wu2024deepseek} and Qwen3-VL~\cite{bai2025qwen3} typically use frozen visual encoders and trainable projectors to map visual inputs to LLM-compatible representations.
Current research focuses on two key areas to improve multi-modal integration:
optimizing training strategies~\cite{bai2023qwen, chen2023minigpt, bai2025qwen25vl, bao2022vlmo} using parameter-efficient techniques, and enhancing visual components by scaling instruction-tuning datasets~\cite{liu2023aligning, zhang2023llavar}.
However, challenges remain in scaling LVLMs, such as limited task generalization and rising inference costs.
To address these issues, the MoE architecture offers an effective solution, enabling efficient parameter scaling while preserving computational efficiency.

\subsection{Mixture-of-Experts}

The MoE architecture~\cite{jacobs1991moe, shazeer2017outrageously, komatsuzaki2022sparse, puigcerver2023sparse, shen2023mixture} improves the efficiency of large-scale models by selectively activating a subset of experts for each token~\cite{wei2024skyworkmoe, xue2024openmoe, jiang2024mixtral, chen2024internvl, bai2023qwen, zhong2024moextend, chen2023lifelong, li2024unimoe, li2025unimoe}, effectively decoupling model capacity from computational cost~\cite{du2022glam, fedus2022switch}.
Current MoE research primarily concentrates on two main areas:
($i$) Efficient utilization of expert resources.
Load balancing strategies~\cite{aminabadi2022deepspeed, eliseev2023fast, lepikhin2020gshard, cai2025ltdr} and expert load prediction methods~\cite{xue2024moeinfinity, fedus2022switch} aim to ensure uniform and effective use of resources.
($ii$) Efficient learning of specific experts.
This involves reducing the interference caused by differences in data features and gradients on the internal parameter learning of each expert~\cite{chen2023octavius, gou2023mixture, zhou2024exploring, liu2024adamole}.
To combine the strengths of both areas, research on dynamic routing has gained significant attention.

\subsection{Dynamic Routing Mixture-of-Experts}
Compared to static routing, dynamic routing aims to optimize resource allocation during inference, preserving representation capability while reducing computational overhead~\cite{huang2024harder, guo2025dynamic, song2025blockffnendsideaccelerationfriendlymixtureofexperts, wang2025remoe, yue2025adak, jin2025moeplusplus, zeng2024adamoe}. 
Existing methods mainly fall into threshold-based~\cite{huang2024harder, guo2025dynamic, song2025blockffnendsideaccelerationfriendlymixtureofexperts, wang2025remoe, yue2025adak} and pseudo-expert~\cite{jin2025moeplusplus, zeng2024adamoe} strategies. 
Threshold-based strategies dynamically limit expert activation using learnable functions or thresholds~\cite{ guo2025dynamic, yue2025adak, song2025blockffnendsideaccelerationfriendlymixtureofexperts}.
For instance, Top-$p$ routing~\cite{huang2024harder} the minimum number of experts required to reach a learnable cumulative threshold, whereas ReMoE~\cite{wang2025remoe} dynamically filters experts via a ReLU activation on the router outputs.
Pseudo-expert strategies implicitly reduce the computational burden of simpler tokens by introducing null or low-cost virtual experts. 
Specifically, AdaMoE~\cite{zeng2024adamoe} introduces zero experts that output zero, and MoE++~\cite{jin2025moeplusplus} further broadens this approach with copy experts for identity mapping and const experts for dynamic interpolation.
However, they fail to effectively model the balance between model complexity and performance.
In this paper, we frame dynamic routing as a MDL problem, linking this objective to gating entropy via information gain.
By utilizing token-level gating entropy as a proxy, our mechanism adaptively aligns expert activation with information density, ensuring efficient and robust utilization.

\section{Methodology}
\label{sec:method}

\subsection{Mixture-of-Experts}

A MoE layer consists of a $K$-experts ensemble $\mathcal{E} = [E_1, E_2, \dots, E_K]$ and a linear layer router $\mathcal{R}$.
As shown in Eq.~\ref{eq1}, for an input $x \in \mathbb{R}^{D}$, the router $\mathcal{R}$ generates weight logits $\mathcal{R}(x) = \mathbf{W} \cdot x$, which are then normalized.
The matrix $\mathbf{W} \in \mathbb{R}^{K \times D}$ represents the lightweight, trainable parameters for routing, and $\mathcal{R}_{norm}(x)_i$ is the routing score of the input $x$ for the $i$-th expert.
The output of the MoE layer, as defined in Eq.~\ref{eq2}, is computed as a weighted sum of the outputs from the Top-$k$ ($k \le K$, the number of activated experts, fixed in static routing) experts with the highest probabilities.
$E_i(x)$ is the output of the $i$-th expert, and the weight for each expert is determined by its routing score.

\begin{equation}
    \mathcal{R}_{norm}(x)_i = \frac{{e^{\mathcal{R}(x)_i}}}{\sum_{j=1}^K {e^{\mathcal{R}(x)_j}}}
    \label{eq1}
\end{equation}
\begin{equation}
    \mathrm{MoE}(x) = \sum_{i\in \text{Top-\textit{k} ids}} \mathcal{R}_{norm}(x)_i \cdot E_i(x)
    \label{eq2}
\end{equation}

Due to the presence of multiple experts, it is necessary to impose the load balancing constraint on MoE layers.
Traditional methods~\cite{fedus2022switch, lin2024moe} introduce a differentiable load balancing loss into MoE layers to encourage a more balanced allocation of tokens across experts.
As shown in Eq.~\ref{eq3}, $\mathcal{F}_i$ is the fraction of tokens processed by expert $E_i$, while $\mathcal{G}_i$ is the average routing probability of expert $E_i$.

\begin{equation}
    \mathcal{L}_{\text{lb}} = K \cdot \sum_{i=1}^K \mathcal{F}_{i} \cdot \mathcal{G}_{i}
    \label{eq3}
\end{equation}

\begin{figure}[t]
    \centering
    \includegraphics[width=1\linewidth]{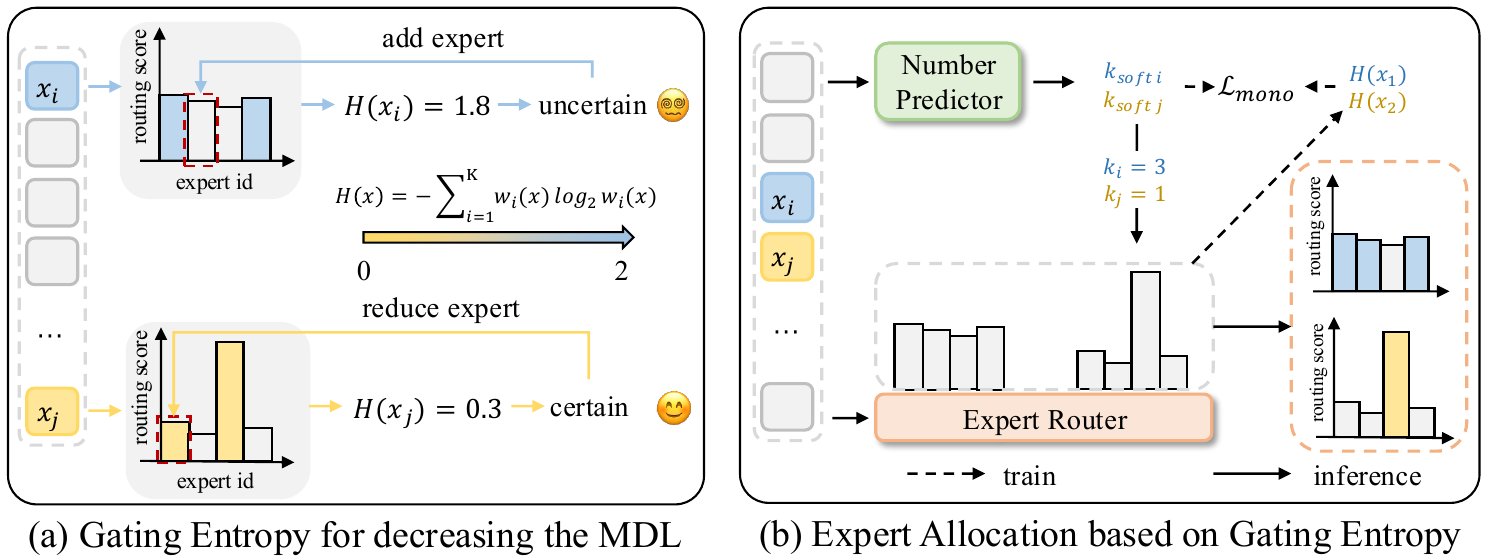}
    \caption{(a). We frame dynamic routing as a MDL problem and demonstrate that gating entropy can serve as a proxy for decreasing the MDL. (b) Based on gating entropy, we model the number of experts required for a token as positively correlated with its gating entropy, allocating more experts to tokens with higher gating entropy.}
    \label{fig_method}
\end{figure}

\subsection{The Overview of Our Method GeMoE}

Token routing is the core mechanism of MoE.
Traditional static Top-$k$ routing leads to inefficient expert utilization.
Although existing dynamic routing methods have made significant improvements, they still fall short of explicitly modeling the trade-off between model complexity and performance.
To address this, we propose a Gating Entropy-based Uncertainty-aware Adaptive Routing for MoE (GeMoE), as illustrated in Fig.~\ref{fig_method}.
Our method consists of two key steps: 
$(i)$ To explicitly model the balance between model complexity and performance, we frame dynamic routing as a MDL problem and demonstrate that gating entropy can serve as a proxy for decreasing the MDL.
$(ii)$ We use gating entropy as prior knowledge to predict the number of expert assignments, thereby establishing a mapping from gating entropy to specific expert allocations.

\subsection{Gating Entropy As a Proxy for Decreasing the MDL}

Token routing can be fundamentally understood as an information encoding process.
Specifically, the token $x$ represents the source information to be encoded.
The expert ensemble $\mathcal{E}$ in MoE can be viewed as a codebook for encoding this source information, where each expert $E_{i}$ corresponds to the $i-th$ codeword in the codebook.
Traditional static Top-$k$ routing can be regarded as a fixed-length encoding scheme, implicitly assuming that the information content of all tokens follows a uniform distribution, which results in the same number of experts being allocated to each token.
This uniform allocation, however, overlooks variations in token complexity, leading to inefficient utilization of expert capacity.

To address this limitation, dynamic routing methods aim to implement a variable-length encoder that adaptively selects the optimal encoding length for each token by choosing the appropriate codewords from the codebook.
This process can be framed as a MDL problem:
minimizing the total encoding length required to describe both the model and its fit to the data, thereby balancing model complexity with performance.
The formalization is as follows:

\begin{equation}
    L(x, \mathcal{E}^\prime) = L(\mathcal{E}^\prime) + L(x \mid \mathcal{E}^\prime), \mathcal{E}^\prime \subseteq \mathcal{E},
    \label{eq4}
\end{equation}
\begin{equation}
    MDL \leftarrow \mathcal{E}^* = \arg\min_{\mathcal{E}^\prime}[L(x, \mathcal{E}^\prime)], \mathcal{E}^\prime \subseteq \mathcal{E}
    \label{eq5}
\end{equation}

\noindent Given a token $x$ and a model $\mathcal{E}'$, where $\mathcal{E}'$ is a subset of the expert ensemble $\mathcal{E}$, the encoding length of the model $L(\mathcal{E}^\prime)$ refers to the number of experts selected.
Longer encoding length indicates higher model complexity.
While $L(x \mid \mathcal{E}^\prime)$ denotes the encoding length of $x$ conditioned on $\mathcal{E}'$, it reflects the model's data-fitting performance based on the selected experts.
Longer length encoding suggests a poorer fit of the model to the data.
Generally, $L(\mathcal{E}^\prime)\uparrow \Rightarrow L(x \mid \mathcal{E}^\prime)\downarrow$.
Since the encoding lengths of both components are model-dependent, determining the subset $\mathcal{E}^*$ that minimizes the encoding length is essential for effectively applying the MDL principle.
However, existing dynamic routing methods rely on heuristic schemes~\cite{jin2025moeplusplus, wang2025remoe, guo2025dynamic, zeng2024adamoe, huang2024harder}, which limits the model’s ability to explore the trade-off between model complexity and performance, as well as its ability to allocate experts effectively.
Fortunately, in MoE-based models, redundancy among experts can occur.
As shown in Eq.~\ref{eq6}, $\mathcal{E}_a^\prime$ is a proper subset of $\mathcal{E}_b^\prime$, and $\mathcal{E}_a^\prime$ may show performance comparable to that of $\mathcal{E}_b^\prime$.
This highlights the potential to reduce model complexity while maintaining performance, specifically by ensuring that $L(\mathcal{E}^\prime)$ decreases without causing an increase in $L(x \mid \mathcal{E}^\prime)$:
$L(\mathcal{E}^\prime)\downarrow \not\Rightarrow L(x \mid \mathcal{E}^\prime)\uparrow$.

\begin{equation}
    L(x \mid \mathcal{E}_a^\prime) - L(x \mid \mathcal{E}_b^\prime) \approx 0, \exists \mathcal{E}_a^\prime \subset \mathcal{E}_b^\prime \subseteq \mathcal{E}
    \label{eq6}
\end{equation}

To enable dynamic routing through MDL minimization, we establish a connection between MDL and information gain.
Suppose the input token to a MoE layer is $x$ and the target output is $y$.
The MDL can be further written as $k \cdot c - \log \sum_{i \in \mathcal{E}^\prime} w_i P_i(y|x)$.
Where $L(\mathcal{E}^\prime) = k \cdot c$, with $k$ denoting the number of experts in $\mathcal{E}^\prime$, and $c$ representing the constant computational penalty for each expert.
$L(x \mid \mathcal{E}^\prime) = - \log \sum_{i \in \mathcal{E}^\prime} w_i P_i(y|x)$, where $w_i$ denotes the routing score and $P_i(y|x)$ denotes the predictive probability of $x$ given by expert $i$.
Our goal is to decrease the MDL.
After adding an expert $E_m$ to the expert set $\mathcal{E}^\prime$, we have a new expert set $\mathcal{E}^\prime_{new}$.
We aim to have $L(x, \mathcal{E}^\prime_{new}) < L(x, \mathcal{E}^\prime)$.
Substituting this into the above formulation yields:

\begin{equation}
    (k+1) \cdot c - \log P(y|x, \mathcal{E}^\prime_{new}) < k \cdot c - \log P(y|x, \mathcal{E}^\prime)
\end{equation}

\noindent Simplifying both sides yields the inequality condition: $\log \frac{P(y|x, \mathcal{E}^\prime_{new})}{P(y|x, \mathcal{E}^\prime)} > c$.
\noindent where we define $\log \frac{P(y|x, \mathcal{E}^\prime_{new})}{P(y|x, \mathcal{E}^\prime)}$ as the information gain. 
If the information gain exceeds a constant $c$, we add the expert; otherwise, we do not.
This serves as the golden rule for dynamic expert allocation.

While information gain offers a rigorous theoretical criterion, directly calculating it requires executing the candidate expert to obtain $P(y|x, E_m)$, which undermines the computational efficiency of dynamic routing.
To address this, we establish a connection between information gain and gating entropy, thereby linking MDL and gating entropy.
Based on the above analysis, when the information gain is large, adding an expert decreases the MDL.
To predict information gain without requiring expert reasoning, we use gating entropy on the token routing distribution as a natural proxy.
Formally, for a token $x$, we employ its routing distribution $\mathcal{R}_{norm}(x) = \{w_1(x), w_2(x), \dots, w_K(x)\}$ based on the router $\mathcal{R}$, and calculate the gating entropy as follows:

\begin{equation}
    H(x \mid \mathcal{R}) = -\sum_{i=1}^{K} w_i(x) \log_2 w_i(x)
    \label{eq8}
\end{equation}

\noindent Higher gating entropy indicates greater routing uncertainty, suggesting a larger potential information gain from adding experts.
After adding an expert $E_m$, $P(y|x, \mathcal{E}^\prime_{new}) \approx P(y|x, \mathcal{E}^\prime) + w_m \cdot P(y|x, E_m)$.
When $w_m$ is close to 0, the information gain $\log \frac{P(y|x, \mathcal{E}^\prime_{new})}{P(y|x, \mathcal{E}^\prime)} \approx \log \frac{P(y|x, \mathcal{E}^\prime) + w_m \cdot P(y|x, E_m)}{P(y|x, \mathcal{E}^\prime)} \approx log(1+\epsilon) \approx \epsilon$, where $\epsilon$ is a small value close to 0.
As the gating entropy increases, the routing weight $w_m$ for newly added expert $E_m$ tends to be higher, resulting in a larger information gain.
For instance, if a token $x$ has $w_1$=0.99 and $w_2$=0.01, adding the expert 2 brings a small information gain $\epsilon$, where the token has the low gating entropy.
Conversely, if $x$ has $w_1$=0.5 and $w_2$=0.5, adding expert 2 brings a large information gain, where the token has the high gating entropy.
Thus, \textbf{when a token has high gating entropy, adding an expert for that token brings the large information gain, thereby decreasing the MDL}.

\begin{figure}[t]
    \centering
    \includegraphics[width=0.9\linewidth]{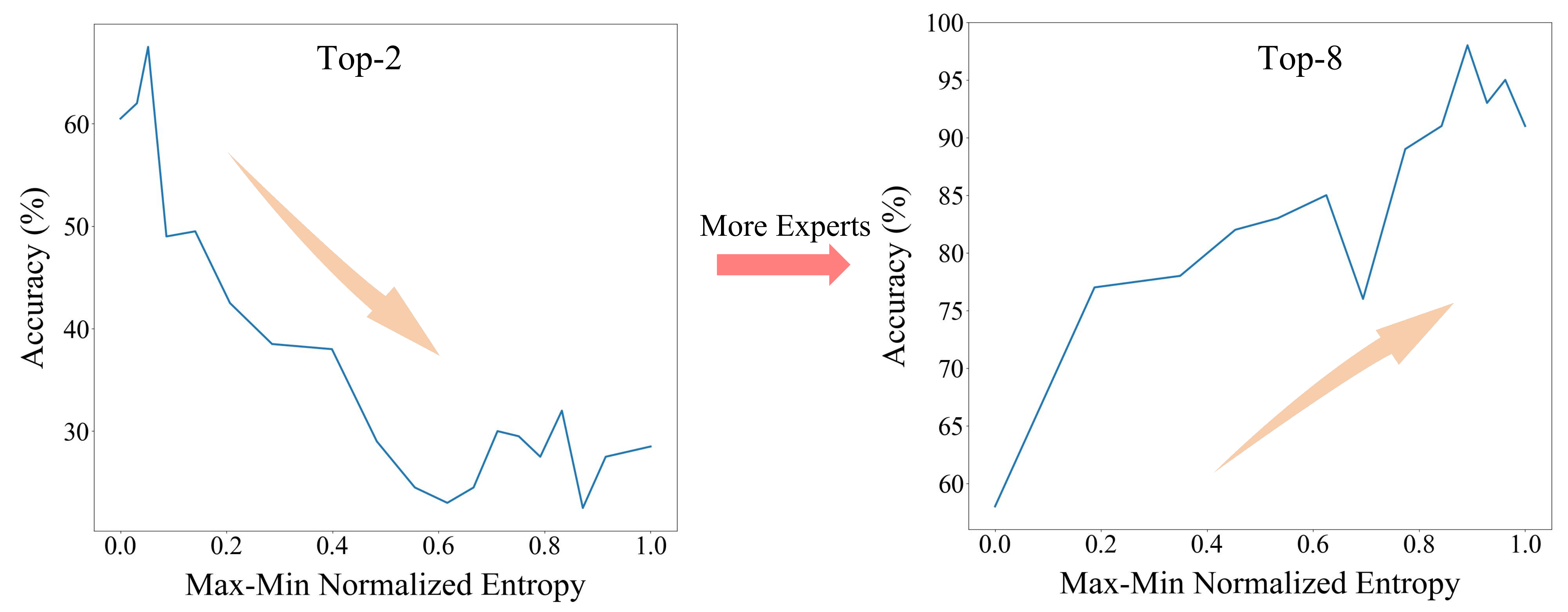}
    \caption{Average accuracy versus maximum-minimum normalized token routing entropy on ScienceQA~\cite{lu2022learn} (4000 samples), evaluated using MolmoE-1B-7B.
    Samples are sorted by normalized token routing entropy and grouped into intervals of 200 samples.}
    \label{fig_analyze}
\end{figure}

To verify the correlation between gating entropy and information gain, we conduct an analysis on the test set.
For each sample, we calculate the model's prediction accuracy at Top-$2$ and Top-$8$, along with the average gating entropy of the tokens in the sample.
We then sort the samples by their gating entropy and group them into fixed-size intervals.
Within each interval, we compute the average accuracy and average gating entropy to analyze potential trends.
As shown in Fig.~\ref{fig_analyze}, the results indicate that increasing the number of experts significantly improves performance for data with high average gating entropy.
However, for data with low average gating entropy, increasing the number of experts does not significantly improve performance and may even result in a negative gain.
These findings clearly demonstrate that adding experts for tokens with high gating entropy brings the large information gain, establishing an explicit relationship between the number of experts and gating entropy.
This enables an efficient dynamic routing mechanism that balances model complexity and performance.

\subsection{Expert Allocation based on Gating Entropy}

Having established the positive correlation between gating entropy and information gain when more experts are added, we aim to assign more experts to tokens with higher gating entropy to help achieve good performance on more uncertain tasks, and fewer experts to tokens with lower gating entropy to effectively reduce the overuse of expert on simple tasks.
This mapping process requires a monotonically increasing relationship between gating entropy and the required number of experts.
To achieve the above goal, we introduce a linear layer expert assignment predictor $\mathcal{P}$ to adaptively learn the mapping process of predicting the number of experts $N=[k_{low}, ..., k_{high}]$ required to be assigned to a token.
$k_{low}$ and $k_{high}$ ($k_{low}=1, k_{high}=8$ in this work.) are the lower and upper limits for expert activation.

\begin{align}
    \label{eq10}
    \mathcal{N}(x)_{i} &= \frac{{e^{\mathcal{P}(x)_i}}}{\sum_{j=1}^{high-low+1} {e^{\mathcal{P}(x)_j}}}, \\
    k_{\text{soft}} &= \sum_{i=k_{low}}^{k_{high}} i \cdot \mathcal{N}(x)_{i-k_{low}} \in [k_{low}, k_{high}], \\
    k &= Round(k_{\text{soft}}) \in N,
\end{align}

\noindent The predictor produce weight logits $\mathcal{P}(x) = \mathbf{W} \cdot x$, where the matrix $\mathbf{W} \in \mathbb{R}^{(k_{high}-k_{low}+1) \times D}$ is the lightweight trainable parameters, and $\mathcal{N}(x)_i $ is the predicting score of the input $x$ to select $N_i$ expert(s).
Given a token $x$, we calculate the probability distribution of the number of experts using $\mathcal{P}$, and then calculate the expected value $k_{\text{soft}}$, which is a soft prediction of the number of experts.
The final $k$ is determined by rounding the $k_{\text{soft}}$ by using a straight-through estimator.

To achieve a monotonically increasing mapping relationship between gating entropy and the number of experts, we employ the monotonic loss.
For any token pair $(x_i, x_j)$ within a batch $X$, if the condition $H(x_i) > H(x_j)$ is satisfied, then $k_{\text{soft}\,i} > k_{\text{soft}\,j}$ is required.
The loss is designed as follows:

\begin{equation}
    \begin{split}
    \mathcal{L}_{\text{mono}} = \sum_{(x_i, x_j) \subseteq X} &
    \begin{cases} 
    \max (0, margin - k_{\text{soft}\,i} + k_{\text{soft}\,j}), & \text{if } H(x_i) > H(x_j) \\
    \max (0, margin - k_{\text{soft}\,j} + k_{\text{soft}\,i}), & \text{if } H(x_i) < H(x_j)
    \end{cases}
    \end{split}
    \label{eq11}
\end{equation}

\noindent Where margin is a boundary threshold used to define the confidence safety boundary that must be reached for a correct prediction.
By combining the cross-entropy loss ($\mathcal{L}_{\text{ce}}$), the proposed monotonic loss $(\mathcal{L}_{\text{mono}})$, and the load balancing loss $(\mathcal{L}_{\text{lb}})$, the overall training objective is formally defined as illustrated in Eq.~\ref{eq12}.
$\alpha$ and $\beta$ are weight hyper-parameters.
This formula simultaneously optimizes the overall balance between task performance, uncertainty-aware expert routing and expert utilization.

\begin{equation}
    \mathcal{L} = \mathcal{L}_{\text{ce}} + \alpha \mathcal{L}_{\text{mono}} + \beta \mathcal{L}_{\text{lb}}
    \label{eq12}
\end{equation}

\section{Experiments}
\label{sec:experiment}

\subsection{Experimental Setup}
\textbf{Backbones.}
Native MoE models, rather than dense-copied versions, serve as backbones, promoting greater diversity among experts and resulting in improved model performance~\cite{radford2019language}.
Thus, we focus on fine-tuning the router to optimize its ability to efficiently route data to the experts.
We use two backbone models with different sizes to validate the effectiveness and generalization of our method: MolmoE-1B-7B (64Top8)~\cite{deitke2025molmo} and DeepSeek-VL2-Tiny-1B-3B (64Top6)~\cite{wu2024deepseek}.

\textbf{Baselines.}
We evaluate four methods of the threshold-based and pseudo-expert dynamic routing strategies.
($i$) DYNMoE~\cite{guo2025dynamic}.
Expert selection occurs only if the similarity exceeds a learned gating threshold.
Additionally, an expert difference loss is introduced to encourage diversity among the experts.
($ii$) Top-$p$~\cite{huang2024harder}.
A learnable module is used to dynamically adjust the expert selection threshold, with a dynamic loss that focuses the probability of expert selection.
($iii$) AdaMoE~\cite{zeng2024adamoe}.
Four null-experts are added to the model, and the router module is trained to decide when to activate them.
($iv$) MoE++~\cite{jin2025moeplusplus}.
Two zero-experts, two copy-experts, and two constant-experts are added, with learnable router parameters.
These baselines offer a comprehensive comparison of various dynamic routing strategies, enabling a fair evaluation of our method.

\textbf{Benchmarks.}
To thoroughly evaluate the vision-language understanding capabilities and the expert activation ratio of each method, we evaluate all methods on comprehensive benchmarks, including MMBench~\cite{liu2023mmbench}, POPE~\cite{li2023evaluating}, ScienceQA~\cite{lu2022learn}, TextVQA~\cite{singh2019towards}, GQA~\cite{hudson2019gqa} and MM-Vet~\cite{yu2023mm}.

\textbf{Configurations.}
To ensure consistency, only the router, along with method-specific trainable modules, are learnable during MoE training across all methods.
All models are trained for one epoch using the LLaVA-1.5-558k~\cite{liu2024visual} dataset.

\definecolor{lightblue}{RGB}{173, 216, 230}
\begin{table}[tb]
    \centering
    \caption{Performance comparison of various routing strategies on the MolmoE-1B-7B~\cite{deitke2025molmo}. $N_{A}$ represents the average number of activated parameters, $Avg_{k}$ is the average number of activated experts, and $Avg_{P}$ is the average performance. \textbf{Bold} highlights the best-performing results, while \underline{underlined} denotes the second-best results. $\uparrow$ and $\downarrow$ indicate that higher or lower values are preferable, respectively.}
    \resizebox{\textwidth}{!}{
        \begin{tabular}{l|c|r|cccccc|c}
        \toprule
        \textbf{Method} & \textbf{N}$_{A}\downarrow$ & \textbf{Avg}$_{k}\downarrow$ & \textbf{MMB} & \textbf{POPE} & \textbf{SQA}$^I$ & \textbf{VQA}$^T$ & \textbf{GQA} & \textbf{MM-Vet} & \textbf{Avg}$_{P}\uparrow$  \\        
        \midrule
        \rowcolor{lightgray!20}
        MolmoE-1B-7B & 1.58B & 8.00 & 58.24 & 85.33 & 87.50 & 56.37 & 50.89 & 45.60 & 63.99 \\
        \midrule
        DYNMoE~\cite{guo2025dynamic} & 1.40B & 6.19 & 51.72 & 87.84 & 81.41 & \textbf{55.16} & \underline{49.95} & 38.90 & 60.83 \\
        Top-$p$~\cite{huang2024harder} & 2.05B & 12.65 & 55.75 & 87.54 & 81.66 & 51.64 & 47.41 & \underline{43.10} & 61.18 \\
        AdaMoE~\cite{zeng2024adamoe} & \underline{1.39B} & \underline{6.13} & \textbf{56.95} & \underline{88.27} & \underline{84.93} & 52.31 & 48.02 & 40.12 & \underline{61.77} \\
        MoE++~\cite{jin2025moeplusplus} & 1.48B & 7.03 & 51.20 & 82.81 & 81.16 & 46.04 & 45.71 & 38.40 & 57.55 \\
        \midrule
        \rowcolor{lightblue!40}
        GeMoE (Ours) & \textbf{1.32B} & \textbf{5.43} & \underline{56.70} & \textbf{88.38} & \textbf{86.98} & \underline{54.88} & \textbf{50.64} & \textbf{44.10} & \textbf{63.61} \\
        \bottomrule
        \end{tabular}
    }
    \label{tab:main}
\end{table}

\begin{figure}[tb]
    \centering
    \begin{minipage}[c]{0.48\linewidth}
        \centering
        \includegraphics[width=1\linewidth]{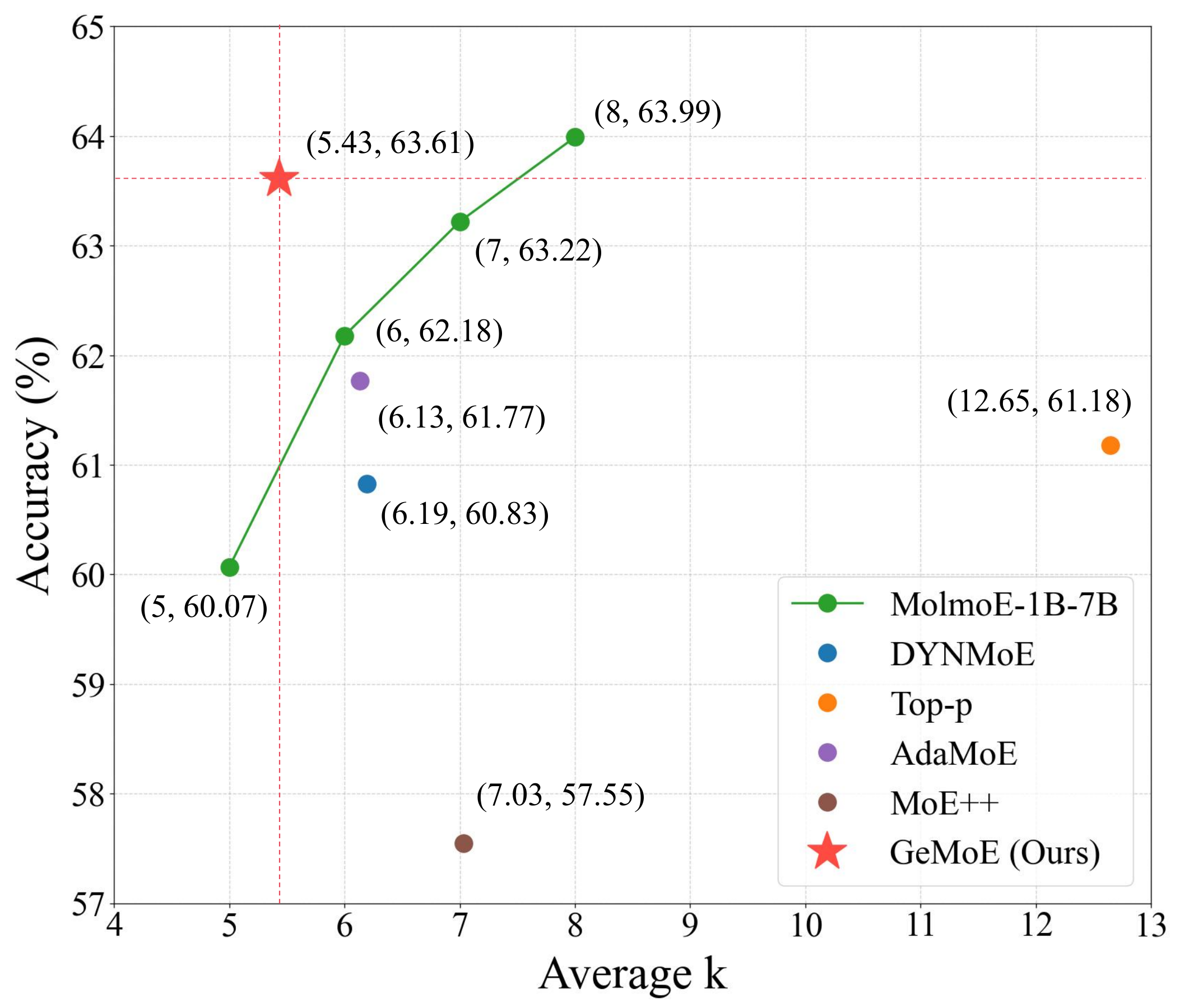}
        \caption{Average accuracy of various Top-$k$ and dynamic routing methods.}
        \label{fig:acc_k}
    \end{minipage}
    \hfill
    \begin{minipage}[c]{0.48\linewidth}
        \centering
        \includegraphics[width=1\linewidth]{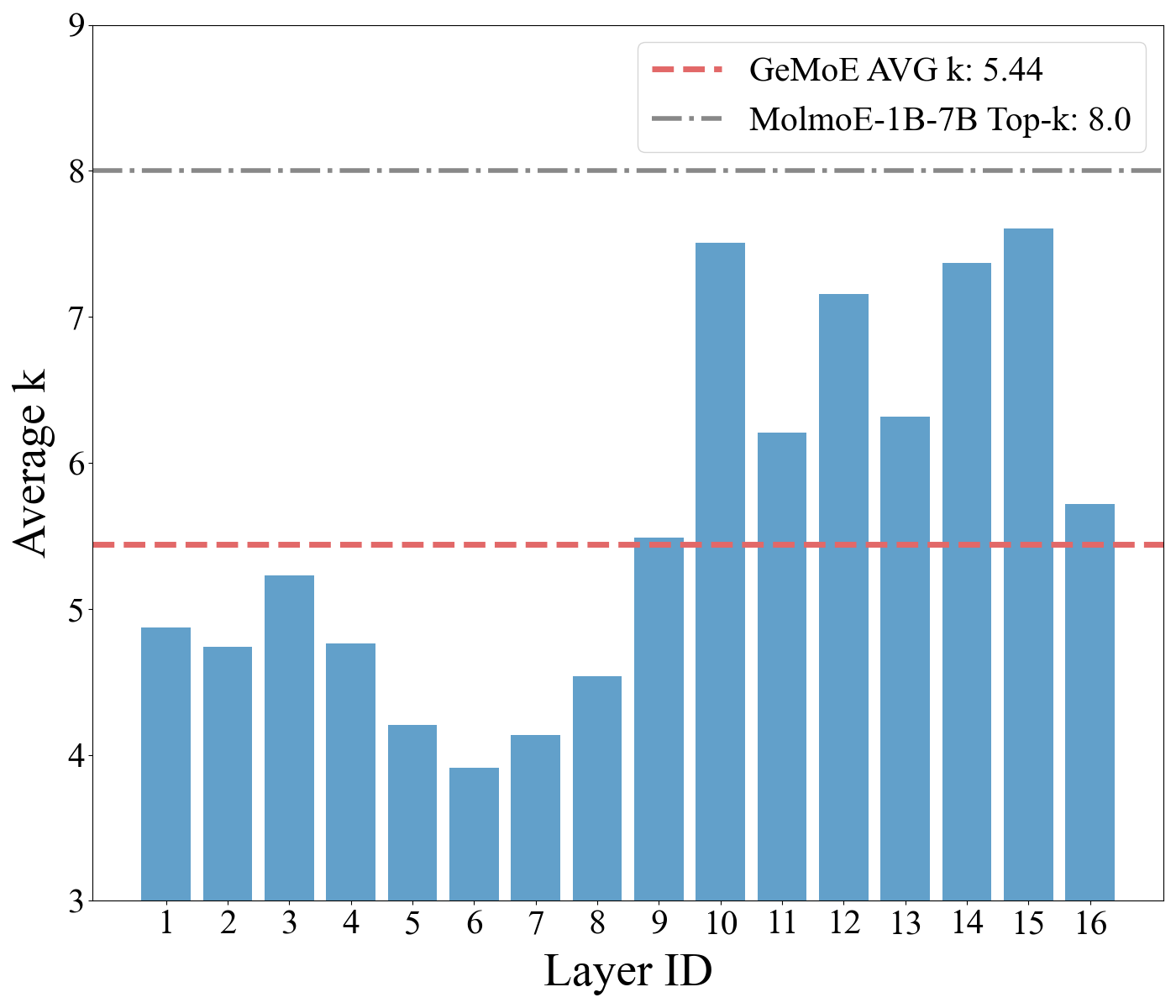}
        \caption{Average number of experts activated in each MoE layer.}
        \label{fig:mean_k}
    \end{minipage}
\end{figure}

\begin{table}[tb]
    \centering
    \caption{Performance comparison of various static Top-$k$ routing on the MolmoE-1B-7B~\cite{deitke2025molmo} and DeepSeekVL2-Tiny-1B-3B~\cite{wu2024deepseek}.}
    \resizebox{1 \textwidth}{!}{
        \begin{tabular}{l|c|c|cccccc|c}
        \toprule
        \textbf{Method} & \textbf{N}$_{A}\downarrow$ & \textbf{Avg}$_{k}\downarrow$ & \textbf{MMB} & \textbf{POPE} & \textbf{SQA}$^I$ & \textbf{VQA}$^T$ & \textbf{GQA} & \textbf{MM-Vet} & \textbf{Avg}$_{P}\uparrow$ \\  
        \midrule
        \multicolumn{10}{c}{\textbf{MolmoE-1B-7B (64Top8)}} \\
        \midrule
        \rowcolor{lightgray!20}
        Top-$8$ & 1.58B & 8.00 & \textbf{58.24} & \underline{85.33} & \textbf{87.50} & \textbf{56.37} & \textbf{50.89} & \textbf{45.60} & \textbf{63.99} \\
        Top-$7$ & 1.48B & 7.00 & \underline{57.30} & 84.79 & 86.51 & \underline{56.07} & \underline{50.76} & 43.90 & 63.22 \\
        Top-$6$ & 1.38B & 6.00 & 53.86 & 84.55 & 85.72 & 55.34 & 50.21 & 43.40 & 62.18 \\
        Top-$5$ & 1.28B & 5.00 & 48.11 & 84.43 & 84.09 & 51.95 & 48.96 & 42.90 & 60.07 \\
        \midrule
        \rowcolor{lightblue!40}
        GeMoE (Ours) & 1.32B & 5.43 & 56.70 & \textbf{88.38} & \underline{86.98} & 54.88 & 50.64 & \underline{44.10} & \underline{63.61} \\
        \midrule
        \multicolumn{10}{c}{\textbf{DeepSeek-VL2-Tiny-1B-3B (64Top6)}} \\
        \midrule
        \rowcolor{lightgray!20}
        Top-$6$ & 1.17B & 6.00 & \textbf{69.56} & \underline{89.07} & \textbf{88.99} & \textbf{70.16} & \textbf{60.84} & \textbf{49.10} & \textbf{71.29} \\
        Top-$5$ & 1.13B & 5.00 & \underline{69.39} & 88.90 & 88.55 & \underline{69.82} & 59.76 & 48.60 & 70.84 \\
        Top-$4$ & 1.09B & 4.00 & 68.87 & 88.70 & 87.24 & 68.72 & 59.63 & 48.50 & 70.28 \\
        Top-$3$ & 1.06B & 3.00 & 67.75 & 88.39 & 87.10 & 67.75 & 59.12 & 47.10 & 69.54 \\
        \midrule
        \rowcolor{lightblue!40}
        GeMoE (Ours) & 1.08B & 3.55 & 68.72 & \textbf{89.41} & \underline{88.78} & 69.24 & \underline{60.37} & \underline{48.89} & \underline{70.90} \\
        \bottomrule
        \end{tabular}
    }
    \label{tab:topk}
\end{table}

\subsection{Main Results}
\textbf{Comparison with State-of-the-Art Dynamic Routing Strategies.}
Tab.~\ref{tab:main} and Fig.~\ref{fig:acc_k} present a comparative analysis of our gating entropy-based uncertainty-aware dynamic routing (GeMoE) against several established dynamic routing baselines on the MolmoE-1B-7B backbone.
The performance of existing dynamic routing strategies is subpar in the native MoE model.
Methods such as DYNMoE, AdaMoE, and MoE++ exhibit significant performance gaps when compared to Top-$8$, despite using an average of 6.13 to 7.03 experts.
On the other hand, Top-$p$ suffers from a notable performance decline, even with an average of 12.65 experts.
In contrast, our approach, which uses the fewest experts on average (5.43), results in only a 0.38\% performance drop compared to Top-$8$, while consistently ranking in the Top-$2$ for performance across all datasets.
This suggests that in MoE models, more experts do not necessarily lead to better performance.
An excessive number of noisy experts can hinder model performance, while the optimal number of experts not only preserves but also enhances expert utilization.
This finding further supports the conclusions drawn in Section.~\ref{sec:method}.

\textbf{Comparison with Various Static Top-$k$ Routing.}
We compare our method against various static Top-$k$ routing strategies on two backbone models of different scales: MolmoE-1B-7B and DeepSeek-VL2-Tiny-1B-3B.
As shown in Tab.~\ref{tab:topk} and Fig.~\ref{fig:acc_k}, our method consistently demonstrates strong model performance and higher sparsity across both backbone models.
Specifically, our approach ranks second only to the top-$8$, with only a 0.38\% and 0.39\% performance drop, despite using 32.1\% and 40.8\% fewer experts compared to Top-$8$, respectively.
This result indicates that while Top-$8$ achieves the highest performance, it suffers from low expert utilization and fails to strike an optimal balance between model complexity and performance.
In contrast, GeMoE offers a better trade-off, maintaining high performance while improving expert utilization and the efficient sparsity of the MoE model.
Additionally, the performance of MolmoE-1B-7B is lower than that of DeepSeek-VL2-Tiny-1B-3B.
We attribute this discrepancy to differences in the distribution of training data.
Specifically, we assume that the experts in the native MoE model already demonstrate strong performance, so we do not fine-tune them.
However, the pretraining data distributions for MolmoE-1B-7B and DeepSeek-VL2-Tiny-1B-3B may differ, with the latter's distribution being closer to that of the test set.

\begin{table}[tb]
    \centering
    \caption{Efficiency comparison of various routing strategies on NVIDIA A800-40G.}
    \resizebox{1 \textwidth}{!}{
        \begin{tabular}{l|r|ccccc}
        \toprule
        \multirow{2}{*}{\textbf{Method}} & \multirow{2}{*}{\textbf{Avg}$_{k}\downarrow$} & \textbf{Param$\downarrow$} & \textbf{Memory$\downarrow$} & \textbf{Inference FLOPs$\downarrow$} & \textbf{Throughput$\uparrow$} & \textbf{Wall-clock Time$\downarrow$} \\
        & & (B) & (GB) & (GFLOPs/token) & (token / second) & (second /sample) \\
        \midrule
        \rowcolor{lightgray!20}
        MolmoE-1B-7B & 8.00 & 7.05 & 32.95 & 80.65 & 1709 & 0.169 \\
        DYNMoE & 6.19 & 7.05 & 32.95 & 72.58 & 1757 & 0.162 \\
        Top-$p$ & 12.65 & 7.05 & 32.95 & 98.44 & 1643 & 0.178 \\
        AdaMoE & 6.13 & 7.05 & 32.95 & 71.22 & 1786 & 0.161 \\
        MoE++ & 7.03 & 7.05 & 32.95 & 75.45 & 1741 & 0.165 \\
        \rowcolor{lightblue!40}
        GeMoE (Ours) & 5.43 & 7.05 & 32.95 & \textbf{68.45} & \textbf{1828} & \textbf{0.152} \\
        \bottomrule
        \end{tabular}
    }
    \label{tab:efficiency}
\end{table}

\textbf{Comparison of Inference Efficiency with State-of-the-Art MoE Routing Strategies.}
We evaluate the inference efficiency of the proposed GeMoE compared to MolmoE-1B-7B and other state-of-the-art routing strategies.
The results are shown in Tab.~\ref{tab:efficiency}.
In terms of memory consumption, all methods use 32.95 GB, the same as MolmoE-1B-7B, indicating that the expert assignment predictor in GeMoE introduces negligible overhead.
GeMoE also delivers significant computational benefits: it reduces inference FLOPs per token by 15.2\% (from 80.65 to 68.45 GFLOPs), leading to a 6.5\% increase in throughput (1828 vs. 1709 tokens/s) and a 10.1\% reduction in wall-clock time per sample (0.152s vs. 0.169s).
Additionally, our method exhibits similar GPU memory usage during training compared to the baselines.
These results demonstrate that GeMoE accelerates MoE inference without increasing memory usage, proving its efficiency.

\subsection{Ablation Studies}

\textbf{Ablation Study of $\mathcal{L}_{mono}$ and Correlation Between Entropy and \textit{k}.}
The core of our method lies in determining the number of experts based on the token-level distribution of gating entropy, ensuring a positive correlation between the number of activated experts and gating entropy.
To validate the effectiveness of this module, we conduct two experiments.
In the first experiment, we remove the gating entropy as a guiding proxy and allow the expert number predictor to learn independently.
In the second experiment, we reverse the constraint, setting a negative correlation instead of a positive one.
As shown in Tab.~\ref{tab:ablation_method}, without the guidance of gating entropy, the predicted number of experts decreased from 5.45 to 4.49, leading to a significant performance drop (from 64.77\% to 60.03\%).
In the second experiment, when the correlation between entropy and \textit{k} becomes negative, the predicted number of experts decreased further, from 5.45 to 2.86, resulting in a greater decline in performance (from 64.77\% to 51.57\%).
The ablation study above validates the positive correlation between gating entropy and the number of experts, demonstrating the effectiveness of the proposed $\mathcal{L}_{mono}$.

\textbf{Comparison of Different Values of $\alpha$ in $\mathcal{L}_{mono}$.}
We conduct a sensitivity study of the weighting $\alpha$ in $\mathcal{L}_{mono}$, as shown in Tab.~\ref{tab:ablation_a}.
Specifically, we evaluate loss weightings of $\alpha \in \{0.0, 0.5, 1.0, 1.5\}$.
The results show that the proposed loss function is robust to different weightings, with the best performance achieved on most datasets when $\alpha = 1.0$.

\begin{table}[tb]
    \centering
    \begin{minipage}[t]{0.49\textwidth}
        \centering
        \caption{Ablation study of $\mathcal{L}_{mono}$ and correlation between entropy and $\textit{k}$. $+/-$ indicate positive/negative correlation.}
        \resizebox{1 \textwidth}{!}{
            \begin{tabular}{cc|c|ccc|c}
            \toprule
            \textbf{$\mathcal{L}_{mono}$} & \textbf{Corr} & \textbf{Avg}$_{k}\downarrow$ & \textbf{MMB} & \textbf{SQA$^\text{I}$} & \textbf{GQA} & \textbf{Avg}$_{P}\uparrow$ \\
            \midrule
             &  & 4.49 & 52.83 & 80.91 & 46.36 & 60.03\\
             $\checkmark$ & $-$ & 2.86 & 41.92 & 72.24 & 40.54 & 51.57 \\
             \rowcolor{lightblue!40}
             $\checkmark$ & $+$ & 5.45 & \textbf{56.70} & \textbf{86.98} & \textbf{50.64} & \textbf{64.77} \\
            \bottomrule
            \end{tabular}
        }
        \label{tab:ablation_method}
    \end{minipage}
    \hfill
    \begin{minipage}[t]{0.49\textwidth}
        \centering
        \caption{Comparison of different values of $\alpha$ used in $\mathcal{L}_{\text{mono}}$ on the MolmoE-1B-7B.}
        \resizebox{1 \textwidth}{!}{
            \begin{tabular}{c|c|cccc|c}
            \toprule
            \textbf{$\alpha$} & \textbf{Avg}$_{k}\downarrow$ & \textbf{MMB} & \textbf{POPE} & \textbf{SQA$^\text{I}$} & \textbf{VQA$^\text{T}$} & \textbf{Avg}$_{P}\uparrow$  \\
            \midrule
            0.0 & 4.49 & 52.83 & 81.77 & 80.91 & 47.65 & 65.79 \\
            0.5 & 5.56 & 56.36 & 86.36 & 85.02 & 54.34 & 70.52 \\
            \rowcolor{lightblue!40}
            1.0 & 5.44 & \textbf{56.70} & \textbf{88.38} & 
            \textbf{86.98} & \textbf{54.88} & \textbf{71.74} \\
            1.5 & 5.45 & 56.42 & 87.87 & 86.12 & 54.68 & 71.27 \\
            \bottomrule
            \end{tabular}
        }
        \label{tab:ablation_a}
    \end{minipage}
\end{table}

\textbf{Average Number of Experts Activated in Each MoE Layer.}
As shown in Fig.~\ref{fig:mean_k}, GeMoE exhibits a variation in the average number of activated experts across different MoE layers.
In general, fewer experts are activated in the shallow layers, while more experts are activated in the deeper layers. 
This indicates that shallow layers deal with relatively simple semantics, requiring fewer experts, whereas deeper layers, handling more complex semantics, need more experts to effectively process the information.
Moreover, the number of expert activations in each layer remains consistently below 8, which helps reduce the computational load.
This results in an average 32\% reduction in the number of activated experts.

\begin{figure}[tb]
    \centering
    \includegraphics[width=1\textwidth]{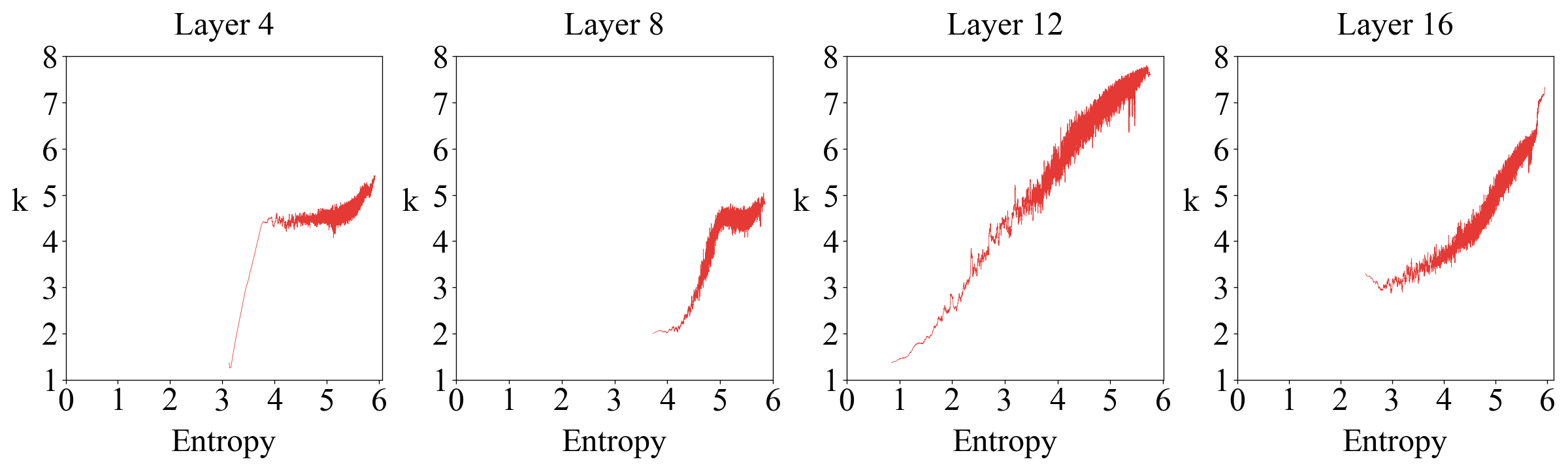}
    \caption{Visualization of the correlation between gating entropy and the number of activated experts on layer 4, 8, 12 and 16.}
    \label{fig:entropy_k}
\end{figure}

\textbf{Correlation between Gating Entropy and the Number of Activated Experts.}
We visualize the correlation between gating entropy and the number of activated experts learned by the expert assignment predictor in layers 4, 8, 12, and 16, with intervals of 4 layers.
As shown in Fig.~\ref{fig:entropy_k}, the gating entropy across different layers leads to varying numbers of activated experts, but all layers exhibit a consistent positive correlation.
Specifically, layers 4 and 8 have higher initial gating entropy compared to layers 12 and 16.
When the gating entropy is small, the number of experts increases rapidly with increasing gating entropy, indicating that gating entropy has a significant impact on expert activation.
However, when the gating entropy reaches a certain value, the increase in the number of experts gradually weakens.
For layers 12 and 16, which have lower initial gating entropy, the increase in the number of experts becomes more steady.
Overall, shallow layers activate fewer experts on average, while deeper layers activate more, consistent with the observations in Fig.~\ref{fig:mean_k}.

\begin{figure}[tb]
    \centering
    \includegraphics[width=1\textwidth]{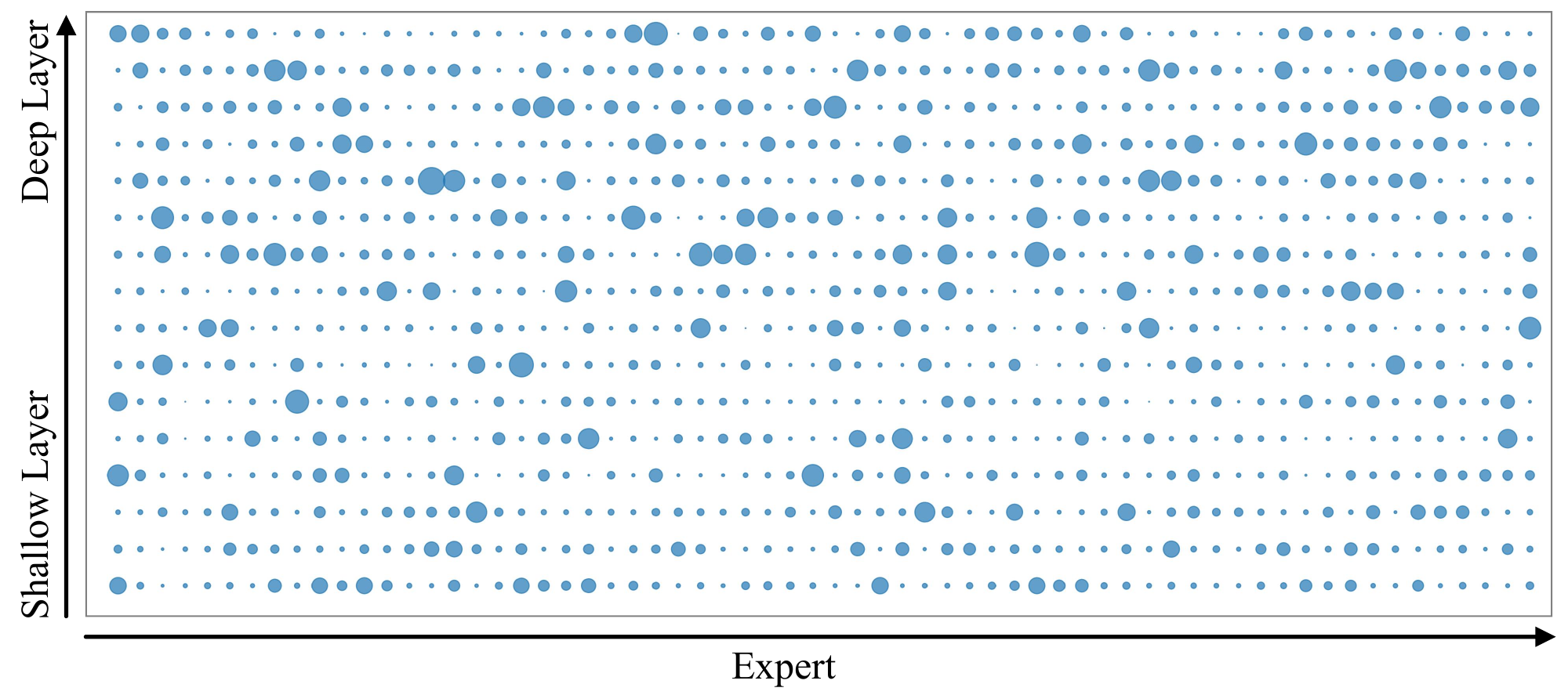}
    \caption{Average number of tokens activated by experts. Larger circle, more activations.}
    \label{fig:expert_act}
\end{figure}

\textbf{Average Number of Tokens Activated by Experts.}
Fig.~\ref{fig:expert_act}  shows the average number of expert activations.
As observed, there is a variation in the number of expert activations across different MoE layers.
Specifically, fewer experts are activated in the shallow layers, while more experts are activated in the deeper layers.
This observation aligns with the findings in Fig.~\ref{fig:mean_k} and Fig.~\ref{fig:entropy_k}, suggesting that shallow layers deal with simpler semantics, requiring fewer experts, while deeper layers handle more complex semantics, necessitating more experts.


\section{Conclusion}
\label{sec:conclusion}

Our study reveals that dynamic routing in MoE fundamentally addresses a Minimum Description Length (MDL) problem by balancing model complexity and performance.
To enable dynamic routing via MDL minimization, we establish a connection between MDL and the gating entropy of token routing.
Specifically, when a token has high gating entropy, increasing the number of experts results in significant information gain, thereby reducing MDL, and vice versa.
Based on this insight, we introduce a learnable, lightweight expert assignment predictor that assigns an adaptive number of experts to each token according to its gating entropy.
To ensure a positive correlation between gating entropy and the number of experts, we incorporate a monotonicity loss.
Through comparisons with various dynamic routing strategies across different backbone models and ablation studies, we demonstrate the effectiveness of GeMoE.
Our method achieves a favorable balance between model complexity and performance while improving the efficiency of expert utilization.

\section*{Acknowledgements}
This work is supported in part by National Science Foundation for Distinguished Young Scholars under Grant 62225605, Zhejiang Provincial Natural Science Foundation of China under Grant LD24F020016, Ningbo Science and Technology Special Projects under Grant 2025Z028, and the Department of Human Resources and Social Security of Xinjiang Uygur Autonomous Region for the financial support of the 2024 ``Tianchi Talents'' Introduction Program (Distinguished Expert Project).

\bibliographystyle{splncs04}
\bibliography{main}

\appendix
\label{sec:appendix}

\section{Methodological Details}

To provide a clearer description of our proposed method, we first present a detailed explanation of the symbols mentioned in the Methodology section, as shown in Tab.~\ref{tab:notation}.
Furthermore, to offer a precise understanding of the practical implementation of our approach, we illustrate both the inference and training procedures in the form of pseudocode, as shown in Algorithm.~\ref{pseudocode}.
This pseudocode representation facilitates reproducibility and allows reviewers to follow the step-by-step logic of our method in a clear and structured manner.

\begin{table}[ht]
\centering
\small
\caption{Detailed notation and description of symbols used in the Methodology.}
\begin{tabular}{c|p{10.5cm}}
    \toprule
    \textbf{Notation} & \textbf{Description} \\
    \midrule
    $\mathcal{E}$ & Expert ensemble \\
    $E$ & Expert \\
    $\mathcal{R}$ & Router \\
    $K$ & Total number of experts \\
    $k$ & Activated number of experts \\
    $x$ & Input token \\
    $\mathcal{R}(x)$ & Routing score of token $x$ \\
    $\mathcal{R}_{\text{norm}}(x)$ & Normalized routing score of token $x$ after softmax \\
    $D$ & Feature dimension processed by the model \\
    $\mathbf{W}$ & Lightweight trainable parameters \\
    $\mathcal{F}$ & The fraction of tokens processed by experts \\
    $\mathcal{G}$ & The average routing probability of experts \\
    $L$ & Encoding length \\
    $\mathcal{E}^\prime$ & A subset of $\mathcal{E}$ \\
    $N$ & The set of expert activation numbers \\
    $k_{low}$ & Minimum number of expert activations \\
    $k_{high}$ & Maximum number of expert activations \\
    $\mathcal{P}(x)$ & Score for selecting the number of experts for token $x$ \\
    $\mathcal{N}(x)$ & Normalized score for selecting the number of experts after softmax \\
    $k_{soft}$ & A soft prediction of the number of experts \\
    $H(x)$ & Gating entropy of token $x$ \\
    \bottomrule
\end{tabular}
\label{tab:notation}
\end{table}

\begin{algorithm}[!t]
  \fontsize{8}{10}\selectfont
  \caption{GeMoE: Gating Entropy-based Uncertainty-aware Adaptive Routing for MoE}
  \label{code}
  \renewcommand{\algorithmicrequire}{\textbf{Inputs:}}
  \renewcommand{\algorithmicensure}{\textbf{Outputs:}}
  \begin{algorithmic}
    \REQUIRE A Batch of Tokens $X =\{x_1, x_2, \dots x_N\}$, Expert Router $\mathcal{R}$, Expert Assignment Predictor $\mathcal{P}$, Experts $\mathcal{E}=\{E_1, E_2, ..., E_K\}$.
    \ENSURE A Batch of Tokens $X^{\prime} =\{x^{\prime}_1, x^{\prime}_2, \dots, x^{\prime}_N\}$, $\mathcal{L}_{mono}$ (only for training).
    \FOR{$i \in \{1,2,...,N\}$}
        \STATE $\mathcal{R}_{norm}(x_i)=softmax(\mathcal{R}(x_i))$ (Equation (1)) 
        \STATE $\mathcal{N}(x_i)=softmax(\mathcal{P}(x_i))$ (Equation (9))
        \STATE $k_{soft}=\sum\limits_{j=k_{low}}^{k_{high}}{j \cdot \mathcal{N}(x_i)_{j-k_{low}}}$ (Equation (10))
        \STATE $k=Round(k_{soft})$ (Equation (11))
        \STATE $x^{\prime}_i=\sum\limits_{j \in Top-k \ ids}{\mathcal{R}
        _{norm}(x_i)_j \cdot E_j(x_i)}$ (Equation (2))
    \ENDFOR
    \IF{training}
        \FOR{$i \in \{1, 2, \dots, N\}$}
            \STATE $H(x_i)=-\sum\limits_j^K{(\mathcal{R}_{norm}(x_i)_j \cdot log_2(\mathcal{R}_{norm}(x_i)_j))}$ (Equation (8))
        \ENDFOR
        \FOR{$i,j \in \{1, 2, \dots, N\}, and \ i\ne j$}
            \STATE $\mathcal{L}_{\text{mono}} = \sum_{(x_i, x_j) \subseteq X} 
            \begin{cases}
                \max (0, \text{margin} - k_{\text{soft} \ i} + k_{\text{soft} \ j}), & \text{if } H(x_i) > H(x_j) \\
                \max (0, \text{margin} - k_{\text{soft} \ j} + k_{\text{soft} \ i}), & \text{if } H(x_i) < H(x_j)
            \end{cases}$ (Equation (12))
        \ENDFOR
    \RETURN $X^{\prime}, \ \mathcal{L}_{mono}$
    \ENDIF
    \RETURN $X^{\prime}$
  \end{algorithmic}
  \label{pseudocode}
\end{algorithm}

\section{Experimental Details}

To validate the reliability and effectiveness of the proposed method, we provide the following supplementary information:
the hyper-parameters settings of the method (Section.~\ref{parameters}), additional evaluation results on additional benchmarks (Section.~\ref{add_bench}), and its generalization performance on language model (OLMoE-1B-7B) (Section.~\ref{add_llm}).

\subsection{Hyper-parameters}
\label{parameters}

The hyper-parameter settings of our method are summarized in Tab.~\ref{tab_parameters}.
Among these, \textit{k}$_{high}$ \& \textit{k}$_{low}$, the margin of $\mathcal{L}_{mono}$, and the coefficient of $\mathcal{L}_{mono}$ are specific to our proposed approach.

\begin{table}[!t]
    \caption{Hyper-parameters for training and inference.}
    \label{tab_parameters}
    \centering
    \scriptsize
    \resizebox{1 \textwidth}{!}{
    \begin{tabular}{cccc}
        \toprule
        \textbf{Epoch} & \textbf{Learning Rate} & \textbf{Learning Rate Schedule} & \textbf{Weight Decay} \\
        1 & 1e-4 & Cosine & 0.0 \\
        \cline{1-4} \\
        \textbf{Coefficient of $\mathcal{L}_{lb}$} & \textbf{Text Max Length} & \textbf{Batch Size per GPU} & \textbf{Train Step} \\
        0.001 & 2048 & 16 & 5000 \\
        \cline{1-4} \\
        \textbf{Precision} & \textbf{\textit{k}}$_{high}$ \& \textbf{\textit{k}}$_{low}$ & \textbf{margin} ($\mathcal{L}_{mono}$) & \textbf{Coefficient of $\mathcal{L}_{mono}$} \\
        Fp16 & 8\&1(MolmoE) / 6\&1(DeepSeek) & $1.2\times(H(x_{i})-H(x_{j}))$ & 1.0 \\
        \bottomrule
    \end{tabular}
    }
\end{table}

\subsection{Evaluation on Additional Benchmarks}
\label{add_bench}

As illustrated in Tab.~\ref{tab:bench}, our approach not only exhibits competitive performance on the eight newly introduced datasets but also maintains desirable sparsity in expert activations, outperforming the baseline method.

\definecolor{lightblue}{RGB}{173, 216, 230}
\begin{table}[ht]
    \centering
    \caption{More comparison of various routing strategies on MolmoE-1B-7B.}
    \resizebox{\textwidth}{!}{
        \begin{tabular}{l|c|cccccccc}
        \toprule
        \textbf{Method} & \textbf{Avg}$_{k}\downarrow$ & \textbf{MMMU}~\cite{yue2024mmmu} & \textbf{AI2D}~\cite{kembhavi2016diagram} & \textbf{InfoVQA}~\cite{mathew2022infographicvqa} & \textbf{MMStar}~\cite{chen2024we} & \textbf{RealWorldQA} & \textbf{OCRBench}~\cite{liu2024ocrbench} & \textbf{VQA}$^{v2}$~\cite{goyal2017making} & \textbf{MME}~\cite{fu2026mme} \\
        \midrule
        MolmoE-1B-7B & 8.00 & 31.2 & 67.9 & 54.2 & 41.8 & 60.5 & 53.5 & 72.51 & 1105.27 \\
        \rowcolor{lightblue!40}
        GeMoE & 5.43 & 30.5 & 67.5 & 54.0 & 41.1 & 60.1 & 52.8 & 72.26 & 1280.80 \\
        \bottomrule
        \end{tabular}
    }
    \label{tab:bench}
\end{table}

\subsection{Evaluation on MoE-based Large Models}
\label{add_llm}

Our GeMoE is fundamentally applicable to a wide range of MoE-based models.
To assess its generalization ability, we conduct additional evaluations on large language models (LLMs).
Specifically, as summarized in Tab.~\ref{tab:llm}, we use OLMoE (64Top8)~\cite{muennighoff2024olmoe} as the backbone model and evaluate our method on four widely adopted language task benchmarks: MMLU~\cite{hendrycks2020measuring}, BigBench~\cite{suzgun2023challenging}, GSM-8K~\cite{cobbe2021training}, and HumanEval~\cite{chen2021evaluating}.
The results suggest that simply activating more experts does not guarantee improved model performance.
For example, on HumanEval, reducing the number of active experts can even lead to better outcomes.
Notably, our method, with only 5.53 experts activated on OLMoE, surpasses the performance of the original Top-$8$ configuration (43.58 vs. 38.59).
By enforcing higher expert sparsity while dynamically assigning the most suitable experts to each token, GeMoE effectively enhances model efficiency and overall performance.
These findings confirm the strong generalization capability of our method across LLMs.


\definecolor{lightblue}{RGB}{173, 216, 230}
\begin{table}[ht]
    \centering
    \caption{Performance comparison of static Top-$k$ routing on the OLMoE-1B-7B~\cite{muennighoff2024olmoe}.}
    \resizebox{1 \textwidth}{!}{
        \begin{tabular}{l|c|c|cccc|c}
        \toprule
        \textbf{Method} & \textbf{N}$_{A}\downarrow$ & \textbf{Avg}$_{k}\downarrow$ & MMLU & BigBench & GSM-8K & HumanEval & \textbf{Avg}$_{P}\uparrow$ \\  
        \midrule
        \multicolumn{8}{c}{\textbf{OLMoE-1B-7B (64Top8)}} \\
        \midrule
        \rowcolor{lightgray!20}
        Top-$8$ & 1.58B & 8.00 & \textbf{41.19} & \underline{30.31} & \underline{68.23} & 14.63 & \underline{38.59} \\
        Top-$7$ & 1.48B & 7.00 & \underline{40.30} & 28.78 & \textbf{68.54} & 9.15 & 36.69 \\
        Top-$6$ & 1.38B & 6.00 & 39.82 & 29.40 & 67.70 & 12.20 & 37.28 \\
        Top-$5$ & 1.28B & 5.00 & 37.99 & 28.88 & 64.97 & \underline{18.90} & 37.69 \\
        \midrule
        \rowcolor{lightblue!40}
        GeMoE (Ours) & 1.32B & 5.53 & 38.55 & \textbf{32.32} & 67.85 & \textbf{28.66} & \textbf{43.58} \\
        \bottomrule
        \end{tabular}
    }
    \label{tab:llm}
\end{table}

\section{Visualization and Analysis}

To provide a more intuitive evaluation of our method, we conduct a comprehensive set of visualization experiments and analyses.
These include the following components:
average $k$ across different datasets;
correlation between entropy and $k$ across different datasets;
expert similarity heatmap;
expert routing paths and expert load analysis.
This visualization framework offers an intuitive understanding of GeMoE’s dynamic routing mechanism, expert utilization, and its effectiveness in handling diverse inputs.

\subsection{Average $k$ Values on Additional Datasets}

\begin{figure}[ht]
    \centering
    \includegraphics[width=1\textwidth]{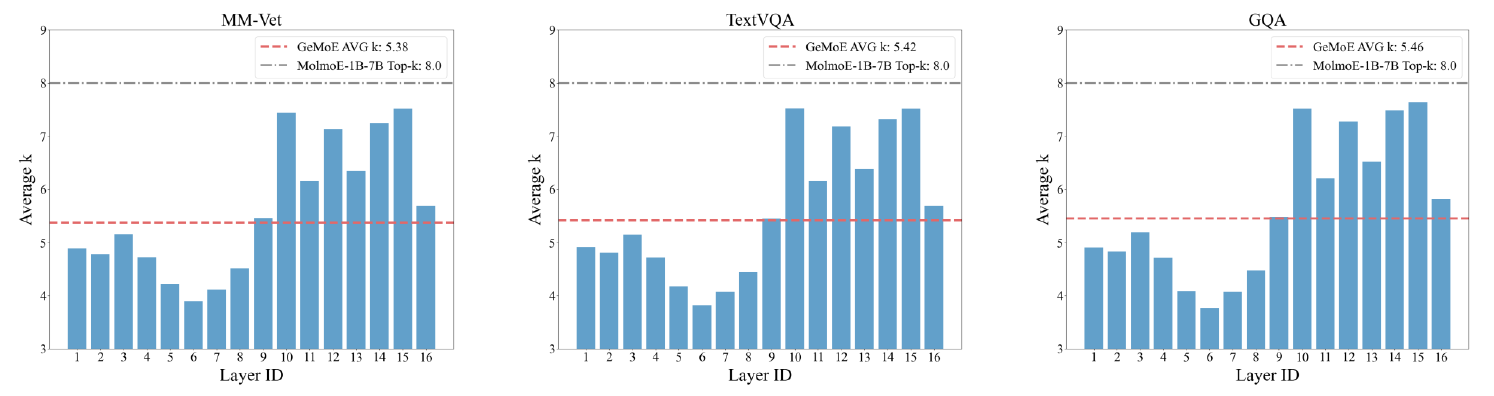}
    \caption{Average number of experts activated in each MoE layer.}
    \label{fig:avg_k}
\end{figure}

As shown in Fig.~\ref{fig:avg_k}, across all three datasets, the average number of experts activated by our method is consistently lower than that of the Top-$8$ baseline.
Notably, fewer experts are activated in the shallow layers, while deeper layers engage more experts.
This design facilitates more efficient expert utilization in simpler semantic spaces, while enabling more thorough information processing in complex semantic spaces.

\subsection{Correlation between Entropy and $k$ on Additional Datasets}

\begin{figure}[ht]
    \centering
    \includegraphics[width=1\textwidth]{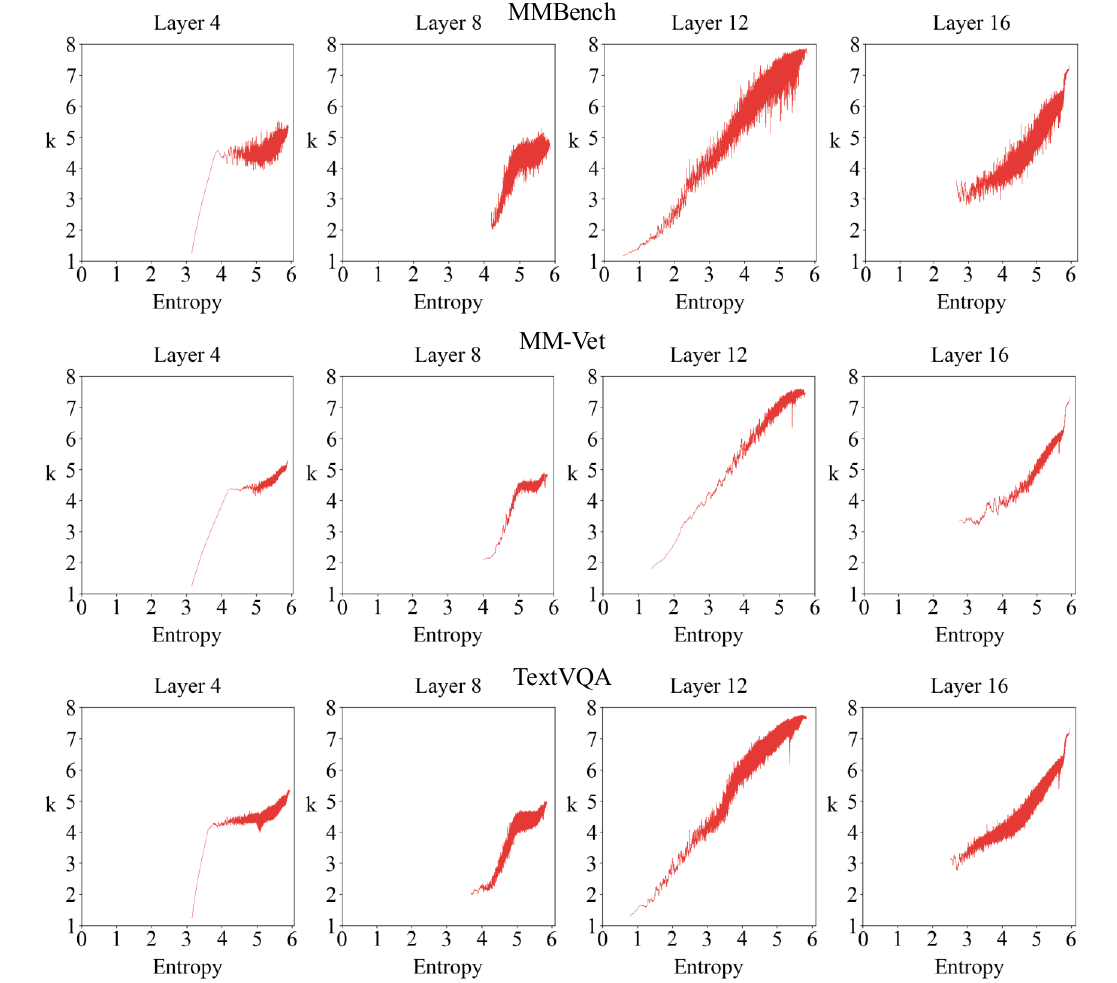}
    \caption{Correlation between gating entropy and the number of activated experts.}
    \label{fig:entropy_k2}
\end{figure}

Across the three datasets as illustrated in Fig.~\ref{fig:entropy_k2}, we observe a consistent positive correlation between entropy and the number of activated experts.
This finding aligns with our proposed hypothesis: when a token has high gating entropy, adding an expert for that token brings the large information gain.

\subsection{Expert Similarity Heatmap}

\begin{figure}[ht]
    \centering
    \includegraphics[width=1\textwidth]{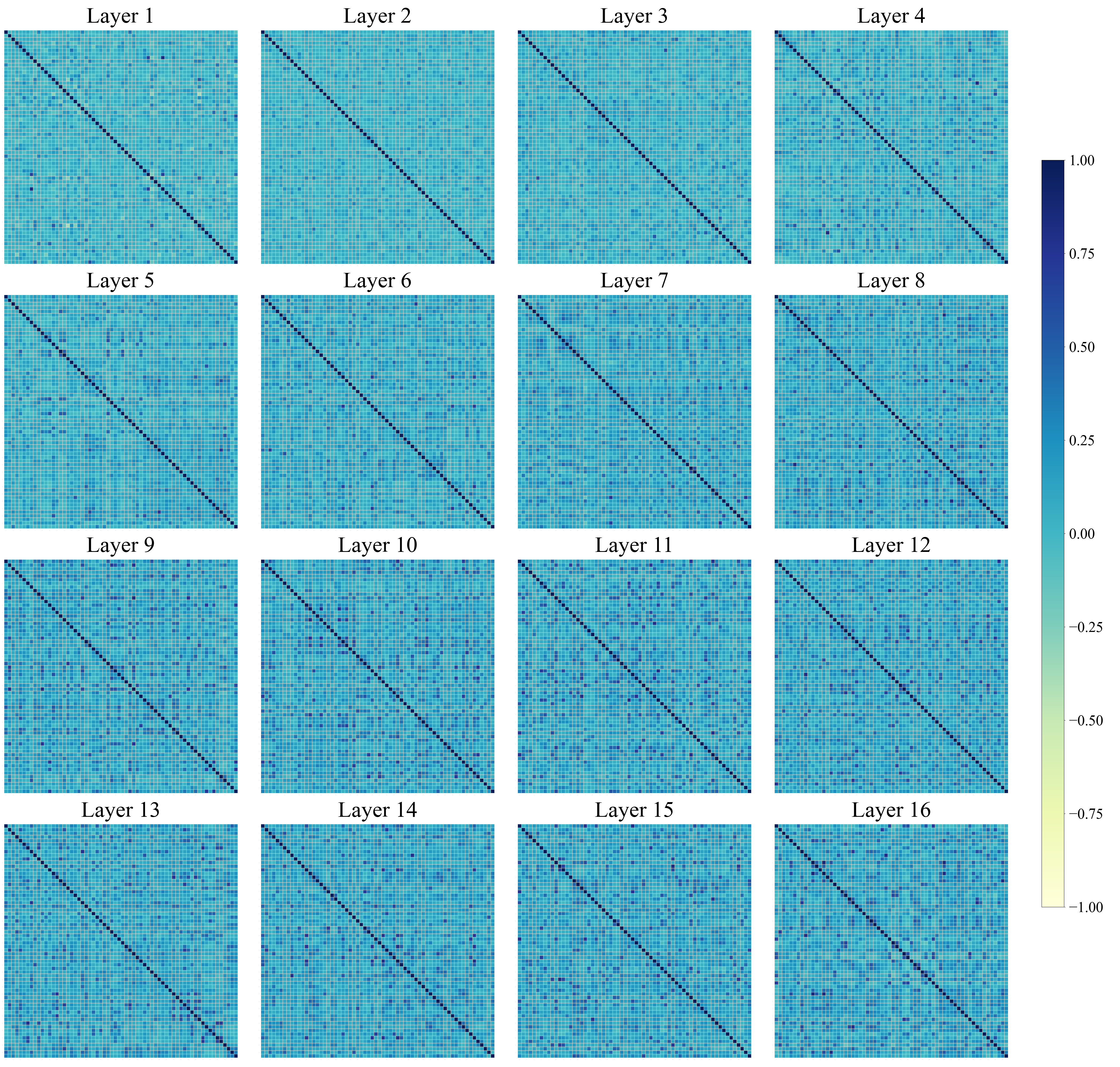}
    \caption{Expert similarity within each MoE layer.}
    \label{fig:expert_sim}
\end{figure}

We further evaluate the diversity among experts by measuring the similarity between their representations in the router.
As illustrated in Fig.~\ref{fig:expert_sim}, the experts learned through our method exhibit clear distinctions.
Such diversity is crucial in MoE architectures, as it promotes the specialization of individual experts and enhances the overall model capacity.

\subsection{Expert Routing Paths}

\begin{figure}[ht]
    \centering
    \includegraphics[width=1.0\textwidth]{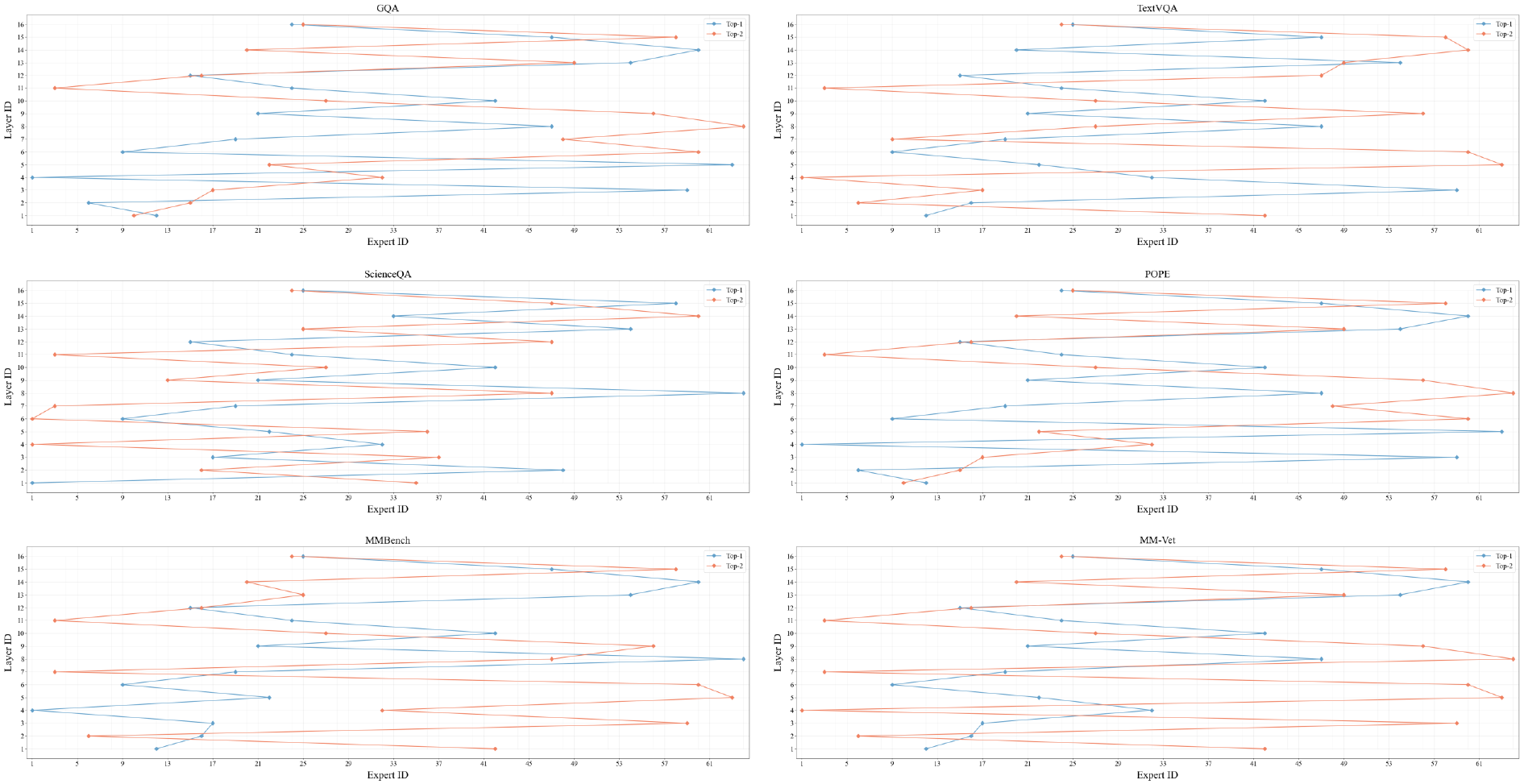}
    \caption{Visualization of activated pathways on Top-$2$.}
    \label{fig:expert_path}
\end{figure}

We present a visualization of the Top-$2$ activated pathways in Fig.~\ref{fig:expert_path}.
It can be observed that the expert activation pathways vary across different datasets, indicating that different data rely on distinct experts within the MoE model.
This highlights the specialization of individual experts and demonstrates their dataset-specific behavior.

\subsection{Expert Load Analysis}

\begin{figure}[ht]
    \centering
    \includegraphics[width=1.0\textwidth]{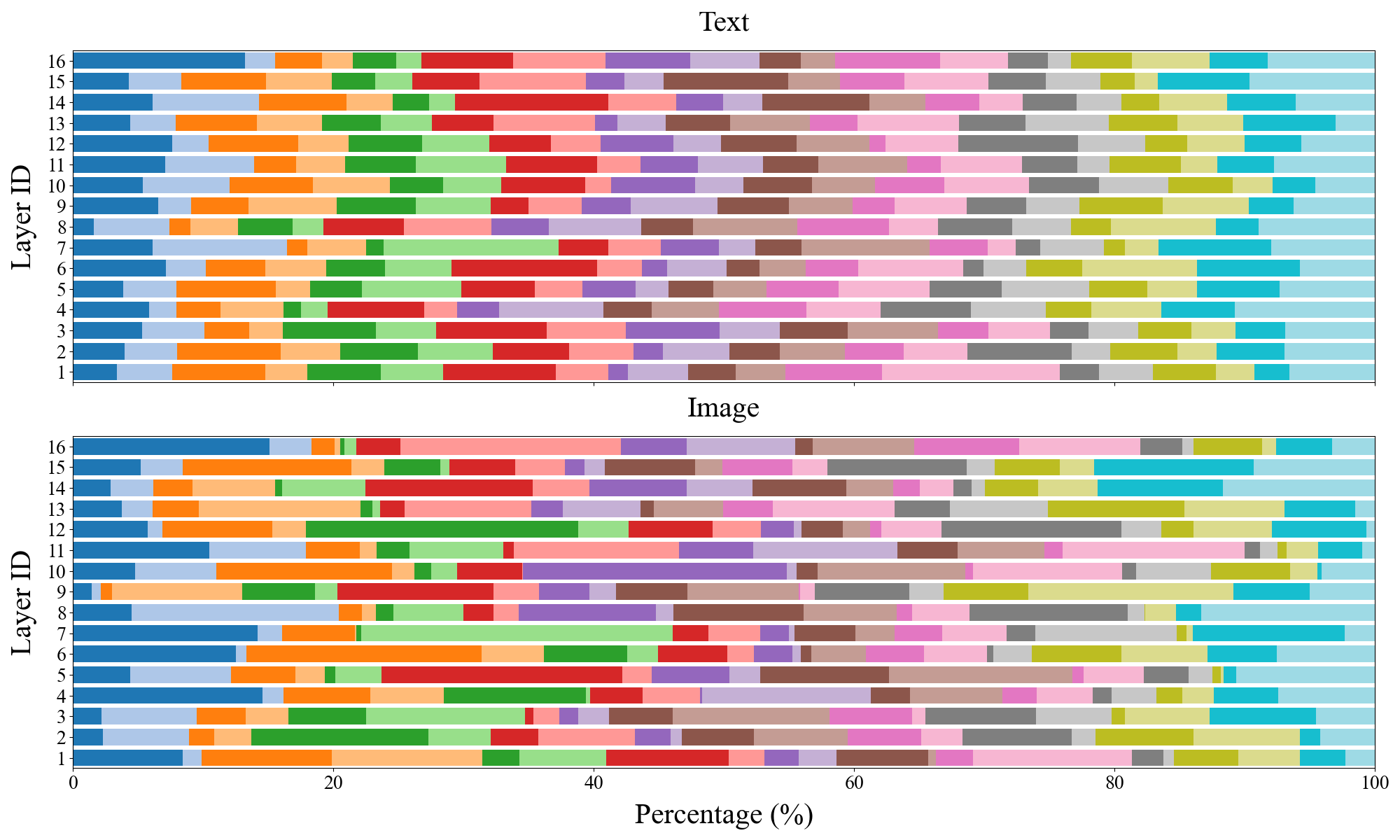}
    \caption{Expert load across different modalities.}
    \label{fig:expert_distribution}
\end{figure}

Fig.~\ref{fig:expert_distribution} illustrates the expert load for text and image modalities.
The load varies across modalities, highlighting the experts’ modality-specific specialization.

\end{document}